\documentclass[runningheads,a4paper]{llncs}

\usepackage{graphicx}
\usepackage[caption=false]{subfig}
\usepackage{amsmath,amssymb}
\usepackage[vlined,algoruled,commentsnumbered]{algorithm2e}
\usepackage{listings}
\usepackage{booktabs}
\usepackage{csquotes}
\usepackage{url}
\usepackage[cleanup={},process=all]{pstool}
\usepackage{anyfontsize}
\usepackage{coverpage}

\title{Path Planning for a Formation of Mobile Robots with Split and Merge}

\author{Estefan\'{\i}a Pereyra$^{1}$, Gast\'on Aragu\'as$^{1}$ and Miroslav Kulich$^{2}$}
\institute{
  $^{1}$Research Centre in Informatics for Engineering, \\
  National Technological University,\\
  C\'ordoba Regional Faculty, Argentina. \\
  \url{{mepereyra, garaguas}@frc.utn.edu.ar}\\
  $^{2}$Czech Institute of Informatics, Robotics, and Cybernetics,\\
  Czech Technical University in Prague, Czech Republic.\\
  \url{kulich@cvut.cz}}

\begin{document}

\mauthor{Estefan\'{\i}a Pereyra, Gast\'on Aragu\'as, and Miroslav Kulich}
\published{{\it Proceedings of the Modelling \& Simulation for Autonomous Systems (MESAS)}, Cham: Springer International Publishing
AG, 2018. p. 59-71. ISBN 978-3-319-76071-1.}
\original{https://link.springer.com/chapter/10.1007/978-3-319-76072-8\_4}
\DOI{10.1007/978-3-319-76072-8\_4}
\coverpage

\maketitle

\begin{abstract}
A novel multi-robot path planning approach is presented in this paper.
Based on the standard Dijkstra, the algorithm looks for the optimal paths for a formation of robots, taking into account the possibility of split and merge.
The algorithm explores a graph representation of the environment,  computing for each node the cost of moving a number of robots and their corresponding paths.
In every node where the formation can split, all the new possible formation subdivisions are taken into account accordingly to their individual costs.
In the same way, in every node where the formation can merge, the algorithm verifies whether the combination is possible and, if possible, computes the new cost.
In order to manage split and merge situations, a set of constrains is applied.
The proposed algorithm is thus deterministic, complete and finds an optimal solution from a source node to all other nodes in the graph.
The presented solution is general enough to be incorporated into high-level tasks as well as it can benefit from state-of-the-art formation motion planning approaches, which can be used for evaluation of edges of an input graph.
The presented experimental results demonstrate ability of the method to find the optimal solution for a formation of robots in environments with various complexity. 

\end{abstract}

\section{Introduction}
\label{sec:intro}
Recent advances in mobile robotics and deployment of robotic systems in many practical applications lead to intensive research of multi-robot systems and robot formations as their special case.
One of the most studied topic deals with planning trajectory/path of a formation in an environment with obstacles, i.e. the problem, how to find a continuous collision-free motion of the formation through a known environment from a current configuration to a given final configuration.
Besides optimization of some parameters of the solution (e.g., path distance, energy consumption, mission time), a shape and size of the formation is constrained and violation of these constrains is penalized.

Approaches to motion and path planning for multi-robot systems and robot formations can be classified into several categories.
Behavior-based algorithms \cite{Balch1998,reynolds1999steering,Pereira2008,Zhong2015} are decentralized and reactive, i.e. each robot is controlled individually, using only local information about its neighborhood.
Robot's behavior is typically composed of several simple behaviors (e.g., separation, alignment, and cohesion in~\cite{reynolds1999steering}), which describe basic actions.
These approaches are easy to implement and applicable to large swarms.
They, on the other hand, fail in finding a plan in complex environments and do not guarantee precise formation control.
To deal with the first problem, several heuristic search based algorithms were introduced based on particle swarm optimization~\cite{Bai2009}, genetic algorithms~\cite{Qu2013509} or ant colony optimization~\cite{Asl2014}.
Nevertheless, precise formation shape can not be still maintained.

In contrast, centralized approaches consider a formation as a single body and plan trajectories in a high-dimensional composite configuration space (CCS).
Exact solutions~\cite{aronov1999motion} are complete, but their complexity is exponential in the dimension of CCS and therefore, methods based on sampling CSS were introduced.
For example, a probabilistic road map with sampling strategies designed especially for multi-robot systems that enable fast coverage of the configuration space was presented in~\cite{Clark2005}, while a generalization of rapidly exploring random trees (RRT) to a graph structure is introduced in~\cite{Kala13}.

Another research stream considers a leader-follower architecture.
Besides other approaches, which compute leader's trajectory and find trajectories of followers relative to this trajectory~\cite{Barfoot2004,Chen2009} or coordinate motion of robots on a preplanned paths~\cite{olmi2008coordination,Liu2011}, a big class of algorithms employs a concept of artificial potential fields.
As classical potential fields~\cite{Zhang2010} tend to find a local optimum, Garrido et al.~\cite{Garrido2011} employed the Voronoi Fast Marching method, which propagates a wave over a viscosity map for a leader.
Trajectories of followers are then dynamically computed to keep desired nominal inter-robot distances using Fast Marching (FM) with incorporated potentials reflecting leader's path, obstacles and positions of other robots.
The method produces paths with minimal Euclidean lengths and avoids local minima, but generated paths are not smooth and go too close to obstacles.
This can be solved by Fast Marching Square (FM$^2$)~\cite{Gomez2013}, which modifies wave expansion by incorporating velocity maps.
Moreover, FM$^2$ manages uncertainties in robot's positions, sensor noise, and moving obstacles.
This was recently applied for formations of unmanned surface vehicles (ships) allowing to model their dynamic behavior~\cite{Liu2015}.
Application of the Frenet-Serret frame to control orientation of a formation enabled path planning for formations of unmanned aerial vehicles in 3D environments~\cite{Alvarez2015a}.

The works mentioned above do not explicitly address splitting and merging of a formation during movement, although this is possible with some approaches.
While centralized exact methods are computationally demanding and thus practically inapplicable for larger formations, random sampling are more promising.
For example, the authors in~\cite{Saska2014deployment} present combination of RRT and particle swarm optimization for cooperative surveillance and demonstrate splitting of a formation in a simplified scenario with a single obstacle.
A reactive obstacle avoidance with added rules for split and merge are introduced in~\cite{Ogren2004}, while extension of flocking behavior~\cite{reynolds1999steering} with game-theoretic based reconfiguration is presented in~\cite{dasgupta2013robust}.

The proposed approach can be seen as a part of a general hierarchical planning algorithm, which consists of a global path planner and a local motion planner.
While the the global planner searches for a topological path of a formation and thus generates primary movement directions, the motion planner determines (based on a path found by the global planner) motion for particular robots in a formation.
This combination prevents the whole planner to be trapped at a local minima and enables to compute feasible trajectories fast.
Similar approach is described in~\cite{Lin2012}, where the global planner constructs a partial Voronoi diagram on the fly and a memetic evolution algorithm is employed for motion generation along this diagram.

The key contribution of the paper lies in design of a novel algorithm for global path planning which extents the well known Dijkstra's algorithm to be applicable for multi-robot systems and which considers possible split and merge of a formation.
This is in contrast with~\cite{Lin2012}, which is primarily focused on a solution of local motion planning, while the global planner is simplified to assume a formation as a single point robot.
On the other hand, motion planning is not addressed in the presented paper as some of the aforementioned approaches can be directly used for it.

The rest of the paper is organized as follows.
The general overview of the approach is described in Section~\ref{sec:approach}, while the proposed extension of Dijkstra's algorithm for robot formations is introduced in Section~\ref{sec:dijkstra}.
In Section~\ref{sec:experiments} we present and discuss some experimental results.
Finally, Section~\ref{sec:conclusion} is dedicated to concluding remarks and future directions of the research.

\section{Framework}
\label{sec:approach}

The planning problem for a fleet of mobile robots can be generally understood as search for a continuous sequence of feasible configurations of the fleet from a start configuration to a given goal configuration.
A feasible configuration is such a configuration, in which a fleet does not collide with a surrounding environment as well as robots in the fleet do not collide with each other.
In many scenarios we don't ask for arbitrary sequence of configurations (trajectory), an optimal trajectory with respect to some criteria (e.g., a mission time) is required instead.
Moreover, additional constrains to fleet's geometry may be applied.
For example, robots may be requested to form and keep a specified shape or lattice or to move close together in order to ensure visibility or communication to their neighbours.

\begin{figure}[hbt]
\centering
\subfloat[][]{
\label{fig:vd-a}
\includegraphics[width=0.485\columnwidth]{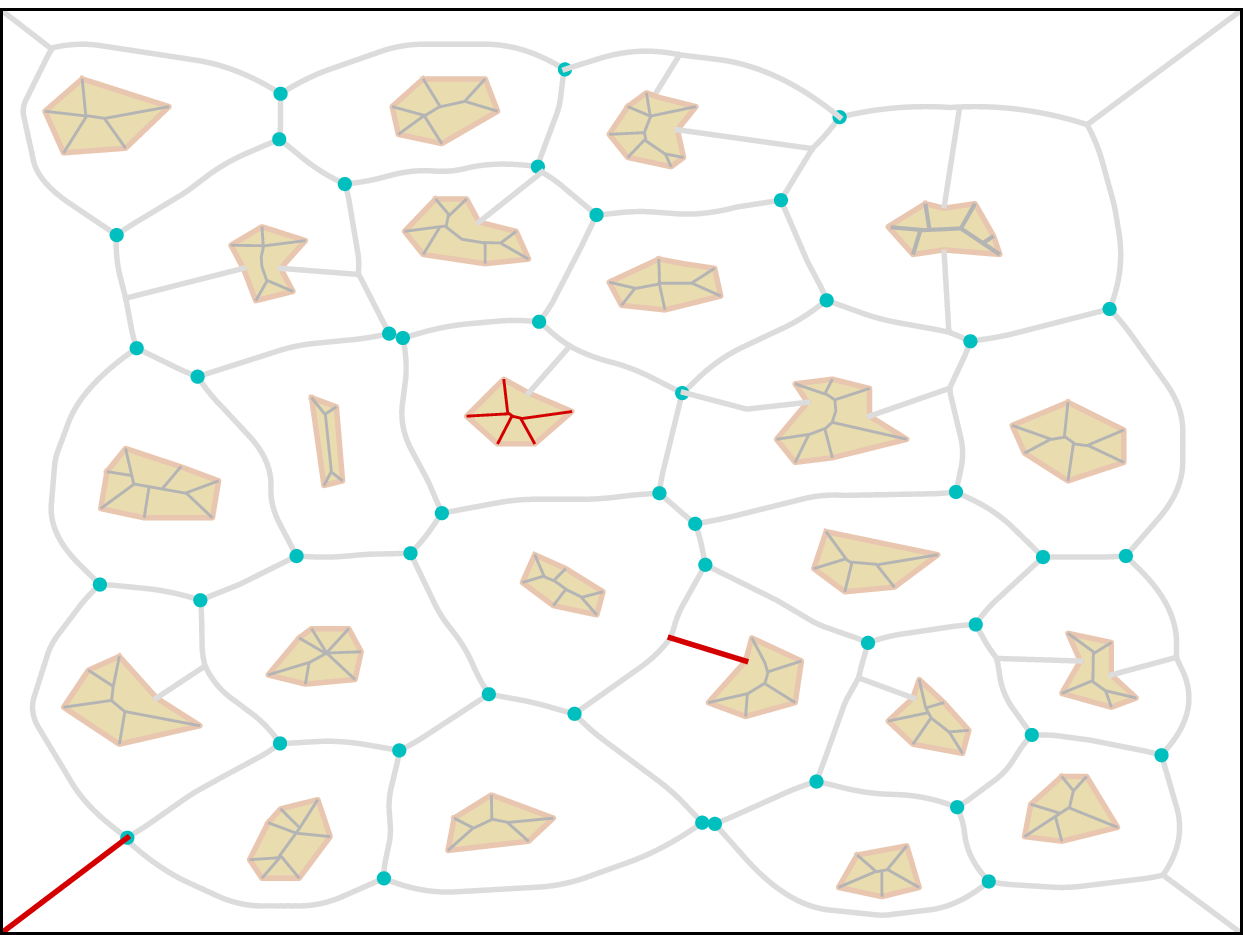}}
\hfill
\subfloat[][]{
\label{fig:vd-b}
\includegraphics[width=0.485\columnwidth]{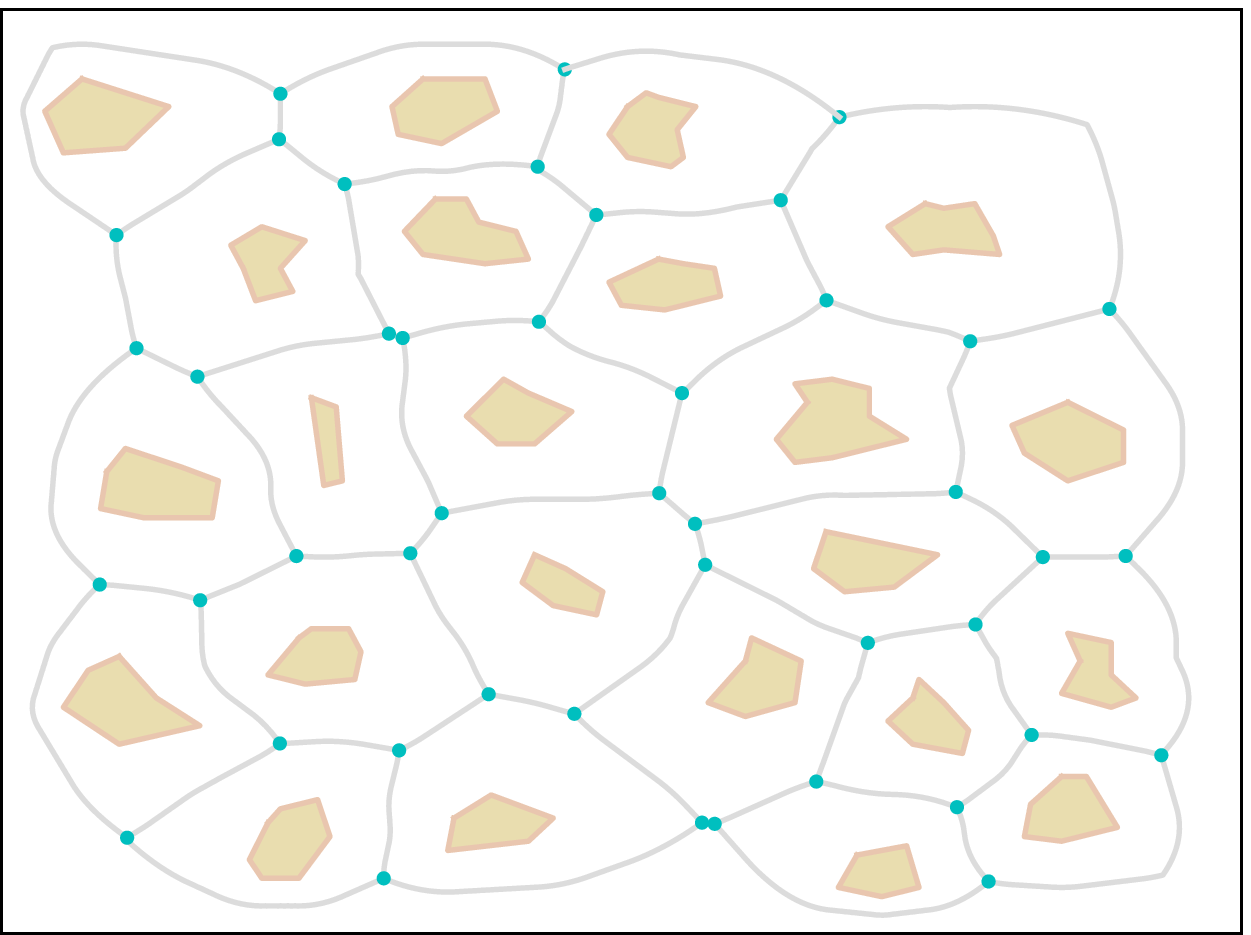}}
\caption[Graph construction.]{Graph construction.
\subref{fig:vd-a} A Voronoi diagram (green) of obstacles (brown) and a rectangular border\footnotemark. The red line represents a tail to be filtered.
\subref{fig:vd-b} The same after filtering of edges and vertices.
}
\label{fig:vd}
\end{figure}

Given a polygonal representation of the environment, start and goal positions, and the number of robots $R$, the proposed solution works in several steps, as described in Algorithm~\ref{alg:framework}.
It starts with construction of a graph $G(V,E)$ (line~\ref{es1}), which is done in three steps.
A Voronoi diagram of obstacles is generated first.
As illustrated in Fig.~\ref{fig:vd}\subref{fig:vd-a}, the original Voronoi diagram contains edges, which are inside obstacles or which are connected to some obstacle vertex.
These edges are thus removed together with nodes and edges forming tails, i.e. components of the graph, which can not be a part of any shortest path except paths originating or ending in these components (one of these components is highlighted red in Fig.~\ref{fig:vd}\subref{fig:vd-a}).
The graph after this removal is shown in Fig.~\ref{fig:vd}\subref{fig:vd-b}.
If the goal or final positions are not in the graph, they are added into it finally.
\LinesNumbered
\begin{algorithm}[ht]
Construct a connected graph $G(V,E)$\;\nllabel{es1}
Evaluate each edge $e\in E$ accordingly to the number of robots\;\nllabel{es2}
For the given start node $Q_{ini}$ compute a shortest path in $G(V,E)$ to each other node $v\in V$\;\nllabel{es3}
Generate motion along the found shortest path to the given goal node(s)\;\nllabel{es4}
\caption{The general planning framework}
\label{alg:framework}
\end{algorithm}

\begin{figure}[!htb]
  \centering 
  \includegraphics[width=0.95\columnwidth]{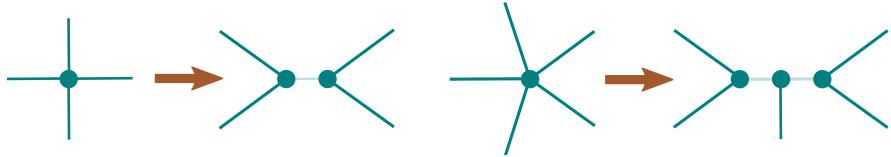}
  \label{fig:split}
  \caption{Substitution of a node with a degree ($\rho$) higher than 3. A node with $\rho=4$ is substituted with two nodes, which are connected with an edge with cost equal to zero (left). Similarly, node with  $\rho=5$ is substituted with three nodes (right). Edges with the zero cost are shown in the light color.}
\end{figure}

All edges of the graph are evaluated in the second step (line~\ref{es2}).
A cost of a particular edges is a vector $\boldsymbol{c}=(c_1,c_2,\dots c_R)$, where $c_k$ is the cost for a formation of $k$ robots to traverse the edge.
The cost of an edge can be either determined as the time needed to traverse the edge with some motion planning algorithm (e.g.~\cite{Gomez2013}) in simulation or it can be approximated based on distance of the edge from the obstacle nearest to it.
Some penalty can be also added to the cost for $k<R$ expressing that the formation is split.
\footnotetext{We use implementation of a Voronoi Diagram (VD) from the Boost Polygon Library (\url{http://www.boost.org/doc/libs/1\_60\_0/libs/polygon/doc/voronoi\_diagram.htm}). The image was generated directly from the output of the library. Note that a Voronoi Diagram for polygon generally contains parabolic segments, which are approximated with line segments in the image.}
For simplicity the algorithm processes all nodes in the graph assuming that they have a degree less or equal to three,  $\rho\leq 3$, that is, they have two or three connected edges.
Nodes with $\rho > 3$ are substituted by an equivalent set of nodes-edges that meet this constraint.
Nodes with $\rho=4$ are substituted with two nodes, which are connected with an edge with cost equal to zero. 
Similarly, nodes with  $\rho=5$ are substituted with three nodes. 
Some examples of this substitution are shown in Fig.~\ref{fig:split}.

The proposed algorithm for formations, described later in Section~\ref{sec:dijkstra}, is run next (line~\ref{es3}), which for a given start computes shortest paths to all other nodes in $G(V,E)$ and all possible sizes of a formation.
In other words, it will be possible to reconstruct an optimal path from the start to an arbitrary node $Q_{goal}$ and the given number of robots $r\in \left<1,R\right>$ after application of this step.
This path is a sequence of nodes $p = \{v_0=v_{ini}, v_1, v_2, \dots, v_n=Q_{goal} \}$ with the properties and constrains detailed in \ref{sec:pathsconstrains}.
A complete motion of a formation is generated finally, given the path $p$ (line~\ref{es4}).
Trajectories of robots in the formation are determined in this step so that relative positions of robots are computed  with respect to geometrical constrains on the formation and to avoid nearby obstacles. 
Again, some motion planning algorithm can be employed to plan a motion between each two consecutive nodes of $p$.

\subsection{Path properties and constrains}
\label{sec:pathsconstrains}
Given a connected graph $\cal G=(V,E)$ representing the working environment, with a set of vertices ${\cal V} = \{v_1, v_2, \dots, v_i\}$ and edges ${\cal E} = \{e_1, e_2, \dots, e_j\}$, and a fleet of robots  ${\cal R} = \{r_1, r_2, \dots, r_R\}$, the corresponding collision-free paths for the fleet in the graph are denoted as ${\cal P} = \{p_1, p_2,\dots, p_R\}$, with $p_i: \mathbb Z^+ \to \cal V$.
Moreover, a path of the individual robot $r_i$ is a sequence of vertices $p_i = \{v_{i_1}, v_{i_2}, \dots v_{i_k}\}$ such that $(v_{i_j}, v_{i_{j+1}})$ is an edge of the graph.
In order to compute the cost of the path another representation of the same path is used, where the number of robots passing through each edge is considered.
In such representation, a path of the individual robot $r_i$ is a sequence of tuples, edges and a number of robots sharing each edge, $p_i = \{(e_{i1}, r_{e_{i1}}), (e_{i2}, r_{e_{i2}}), \dots (e_{ik}, r_{e_{ik}})\}$, where $(e_{ij}, e_{i(j+1)})$ are connected edges of the graph.

The feasibility of the path $p_i$ is conditioned upon the following constrains: 1) initially, all the robots in the fleet are in the start position: $p_i(0) = Q_{\rm ini}, \,\forall p_i \in \cal P$.
2) there exists a state in the path $k_{\rm min}\in \mathbb Z^+$ such that $p_i(k_{\rm min}) = Q_{\rm goal}$, meaning that the robot $r_i$ reaches the goal on the shortest possible path. 
3) any two paths from ${\cal P}$ have not to be in a collision, i.e. given any pair of states $m,l \in \left<0, k_{\rm min}\right>$, two paths $p_i, p_j$ are in collision if $(p_i(m), p_i(m+1)) = p_j(l+1), p_j(l))$.
4) given two paths $p_i, p_j$ and two states $m,l \in \left<0, k_{\rm min}\right>$, if $p_i(m) = p_j(l)$ then $m = l \gets max(m,l)$, i.e. the robot that arrives first to a vertex waits for the second one, in order to keep the formation joint as much time as possible.

\section{Multi-robot path planning algorithm}
\label{sec:dijkstra}

Standard Dijkstra's algorithm finds the cheapest paths together with their costs from the given source node $Q_{\rm ini}$ to all other nodes in the graph ${\cal G}$ given costs of all graph edges. 
The algorithm stores for each node $v\in {\cal G}$ the minimum cost $C_{v_{\rm min}}$ to reach this node and the predecessor $v_{\rm prev}$ from where the node is reached.
The shortest path for the given node $v$ can be then easily determined by walking consecutively over predecessors starting in $v$.
In the initialization stage, costs $C_v$ of all the nodes are set to infinity and their predecessor $v_{\rm prev}$ to a fictive node $None$, what means that the shortest path has not been found yet.
The only exception is the start node, whose cost value is set to $0$.
The start node is then put into a priority queue, which is sorted according to $C_{v_i}$.
The algorithm then consecutively takes the nodes from the priority queue and for the current node $u$, their neighbors $v_i$ are processed.
For each neighbor $v_i$, the total cost $C_{v_i}$ of arriving to it from $u$ is computed.
If $v_i$ is reached for the first time by the algorithm, it is added into the priority queue with its total cost $C_{v_i}$ corresponding of arrive to it from $u$, and $u$ is assigned as its predecessor, $v_{\rm prev} = u$. 
If it is not the case, the computed cost $C_{v_i}$ is compared with the cost stored in the node and, if it is smaller, the staged cost is updated to $C_{v_i}$, $v_{\rm prev}$ is set to $u$, and $v_i$ with its new cost is added to the priority queue.

For the multi-robot case, the cost of an edge can vary with the size of the formation traversing the edge, and thus, all possible combinations of formation sizes traveling through the node must be considered. 
This means that a node can not be processed at once, because it is not guaranteed that all needed information is available until all the robots are processed. 
The proposed algorithm for a formation is depicted in Algorithm~\ref{alg:exhaustive}.
Here, a formation of $R$ robots, a connected graph $\cal G(V,E)$, and a vector of costs ${\boldsymbol c}^e = \left( c^e_1, c^e_2, \dots, c^e_R\right)$ for each edge  $e$, where $c^e_r$ is the cost paid for traversing $e$ with $r \in \left<1,R\right>$ robots are considered.
\begin{algorithm}
  \caption{Multi-robot path planning \label{alg:exhaustive}}
  \KwIn{$\cal G(V,E)$ - connected graph\\
  $R$ - number of robots\\
  $Q_{\rm ini}$ - start node\\}
  \KwOut{$\cal P$ - set of paths}
  \BlankLine
  \hrule
  \BlankLine
  \ForEach{node $v \in \cal G(V,E)$}{\label{ald:init1}
    ${\cal N}.add(v)$ 
  }
  \ForEach{number of robots $r \in \{1, \dots, R\}$}{
    \ForEach{neighbour $v$ of $Q_{\rm ini}$}{
        $c_{\rm vr} \gets compute\_cost(v,r)$\\
        $p_{\rm vr} \gets compute\_path(v,r)$\\
        $s_{\rm vr} = \left< v, r, c_{\rm vr} , p_{\rm vr}\right>$\label{alg:createinist}\\
      ${\cal H}.add(s_{\rm vr})$ \label{alg:addtoqueue}\\
      ${\cal S}.add(s_{\rm vr})$ \label{alg:addtostt}\\
    }
  }
  ${\cal N}.remove(Q_{\rm ini})$\label{ald:endinit}\\
  \While{${\cal N}$}{\label{ald:loop}
  $\left<u, r_{\rm u} ,c_{\rm ur} ,p_{\rm ur}\right>\gets s_{\rm ur} = {\cal H}.pop()$\label{alg:currentSt}\\
  \If{$p_{\rm ur}\notin{\cal P}$}{
    ${\cal P}.add(p_{\rm ur})$\label{alg:addtoP}
  }
  \ForEach{number of robots $r \in \{1, \dots, r_{\rm u}\}$}{
      \ForEach{neighbour $v$ of $u$}{
        $c_{\rm vr} \gets compute\_cost(s_{\rm ur}, v, r)$\\
        $p_{\rm vr} \gets compute\_path(s_{\rm ur}, v, r)$\\
        $s_{\rm vr} = \left< v, r, c_{\rm vr} , p_{\rm vr}\right>$\label{alg:createst}\\
        \If{$s_{\rm vr} \not\in {\cal S}$}{
          $combine\_states(s_{\rm vr},{\cal S})$\\
          ${\cal H}.add(s_{\rm vr})$\\
          ${\cal S}.add(s_{\rm vr})$
        }
      }
    }
  \If{$r_{\rm u} = R$}{
    ${\cal N}.remove(u)$\label{alg:removenode}
  } 
 }
\end{algorithm}
Similarly to the standard Dijkstra's algorithm, the algorithm uses a priority queue $\cal H$ to select the next state to be processed.
In our approach, a state is a data structure associated to a node $v$, given by $s_{\rm vr} = \left<v, r_{\rm v}, c_{\rm vr}, p_{\rm vr}\right>$, where $r_{\rm v}$ is the number of robots arriving to the node, $p_{\rm vr}$ are the paths for each of this robots computed from $Q_{\rm ini}$ to $v$, and $c_{\rm vr}$  is the highest cost corresponding to the worst paths of $p_{\rm vr}$. 
A node can have a lot of associated states for the same number of robots, depending on the combination of paths used for each robot in the node. 
All of these states are stored in a state table $\cal S$ which will be used later to combine paths in the merge operation.

The algorithm starts with initialization of data structures (lines~\ref{ald:init1}~--~\ref{ald:endinit}).
First, all nodes in the graph are stored in a dynamic list $\cal N$.
Then, the states associated to each neighbour of $Q_{\rm ini}$ considering a different number of robots $r\in\left<1,R\right>$ are generated (line~\ref{alg:createinist}) and stored in the queue (line~\ref{alg:addtoqueue}) and in the state table (line~\ref{alg:addtostt}).
After that, the algorithm loops processing each of the states from the priority queue (line~\ref{ald:loop}) until the optimal paths for the $R$ robots from $Q_{\rm ini}$ to each node in the graph are computed.

The algorithm processes each state as follows.
It takes the next state $s_{\rm ur}$ from the queue (line~\ref{alg:currentSt}).
If this state corresponds to an arrival of $r$ robots to the node $u$ from $Q_{\rm ini}$ for the first time, what means that it is the cheapest solution so far, it is stored in the set of optimal paths $P$ (line~\ref{alg:addtoP}).
Then, considering the possibility of formation split, for each number of robots $r < r_{\rm u}$, and for each neighbor $v$ of $u$, a new state $s_{\rm vr}$ is generated (line~\ref{alg:createst}).
If this state does not exist in the state table $\cal S$, it is combined if possible with other states associated with the node $v$ and stored in $\cal S$.
In this step, the possibility of formation merge is considered by means of  combination of states.
Two states are combinable if they meet a set of rules, which will be described in section~\ref{sec:stcombination}.
Finally, if $r_{\rm u} = R$, the node $u$ is removed from the list $\cal N$ (line~\ref{alg:removenode}).

\subsection{Merge of the formation}
\label{sec:stcombination}
The process of states combination allows the algorithm to deal with formation merge.
A state $s_{\rm vk}$ represents the arrival of $k$ robots to the node $v$ by means of their $p_{\rm vk}$ paths.  
The existence of a second state $s_{\rm vq}$ associated to the same node $v$, such that  $k+q <= R$, allows the combination of both states which leads to generation of a new state $s_{\rm v(k+q)}$.
 The new state represents the arrival of $k+q$ robots to the node $v$ by means of a combination, that is a merge of $k+q$ robots of the formation. 
In order to evaluate whether a merge is possible, a set conditions must be satisfied.
The fuction to combine two sattes is depicted in Algorithm~\ref{alg:combinest}. 
Given a state $s_{\rm vr}$, the function tries to combine it with all other states in the state table $S$, ensuring that the number of robots after the combination remains less or equal to $R$ (lines~\ref{alg:combineR} -~\ref{alg:combineS}).
Each possible combination is evaluated by three rules (lines~\ref{alg:edgerule} -~\ref{alg:crossrule}), and if all of them are satisfied the new combined state is added to the queue $\cal H$ and state table $\cal S$ (lines~\ref{alg:evalinit} -~\ref{alg:evalend}).
\begin{algorithm}
  \caption{Function \textit{combine\_states()}\label{alg:combinest}}
  \KwIn{$s_{\rm vr}$ - state\\
  $\cal S$ - states table\\
  $R$ - number of robots}
  \KwOut{New combined states}
  \BlankLine
  \hrule
  \BlankLine
  $p_{\rm vr}, r\gets s_{\rm vr}$\\
  \ForEach{number of robots $i \in \{1, \dots, R - r\}$}{\label{alg:combineR}
    \ForEach{state  $s_{\rm {v}i} \in \cal{S}$}{\label{alg:combineS}
      $p_{\rm {v}i}\gets s_{\rm {v}i}$\\
      $passed = shared\_edges\_rule(p_{\rm {v}i}, p_{\rm vr})$\label{alg:edgerule}\\
      $passed = shared\_nodes\_rule(p_{\rm {v}i}, p_{\rm vr})$\label{alg:noderule}\\
      $passed = cross\_edges\_rule(p_{\rm {v}i}, p_{\rm vr})$\label{alg:crossrule}\\
      \If{$passed$}{ \label{alg:evalinit}
        $s_{\rm v(r+i)} = \left< v, r+i, c_{\rm v(r+i)} , p_{\rm v(r+i)}\right>$\label{alg:createmergest}\\
        ${\cal H}.add(s_{\rm v(r+i)})$\\
        ${\cal S}.add(s_{\rm v(r+i)})$
      } \label{alg:evalend}
    }
  }
\end{algorithm}

Using the edge representation of paths, 
$$p_i = \{(e_{i1}, r_{e_{i1}}),(e_{i2}, r_{e_{i2}}), \dots,(e_{ik}, r_{e_{ik}})\},$$ 
two paths $p_1$ and $p_2$ must fulfill the following rules in order to be combined:\\
\textbf{Shared edges rule}
This rule looks for edges common to both paths, and verifies whether those shared edges contain equal number of robots.\\
\textbf{Shared nodes rule}
This rule looks for nodes common to both paths, and  verifies whether the number of robots arriving to the node is higher than or equal to the number of robots leaving the node.\\
\textbf{Cross edges rule}
This rule avoids combinations where both paths contain the same edge but in opposite direction, given that these combinations don't meet the no collision constrain for paths (Section~\ref{sec:pathsconstrains}).

\section{Experimental evaluation}
\label{sec:experiments}
The proposed algorithm for multi-robot path planning has been implemented in Python 2.7.
All the experiments were evaluated under the same conditions, using a notebook with an \texttt{Intel CORE i5} processor and 4GB of RAM, running a Debian GNU/Linux. 
Given that the algorithm is based in a Voronoi diagram of the map, it is complete; and since it is based on the standard Dijkstra, it is optimal too.
In order to prove its optimality, the experiments were performed on a set of maps with various complexities  from \url{http://imr.ciirc.cvut.cz/planning/maps.xml}.
Nevertheless, because of the high complexity of the problem, the optimality of the solutions was possible to verify manually only for maps with a low number of nodes and with a low number of robots.

\begin{figure}[htb]
  \centering
  \subfloat[]{ 
    \label{subfig:simple_map_a}
    \includegraphics[width=0.48\textwidth]{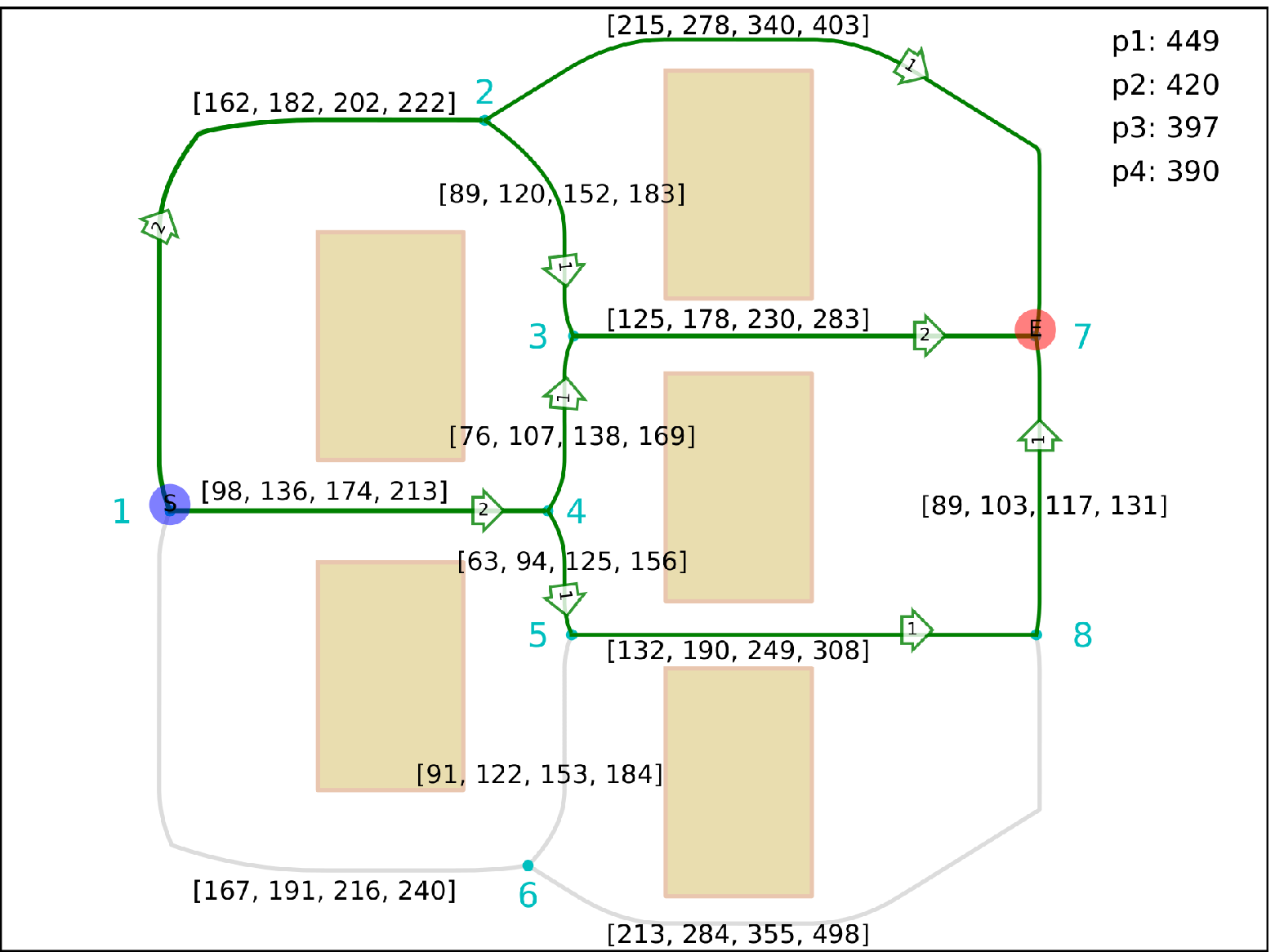}
  }\hfill
  \subfloat[]{ 
    \label{subfig:simple_map_b}
    \includegraphics[width=0.48\textwidth]{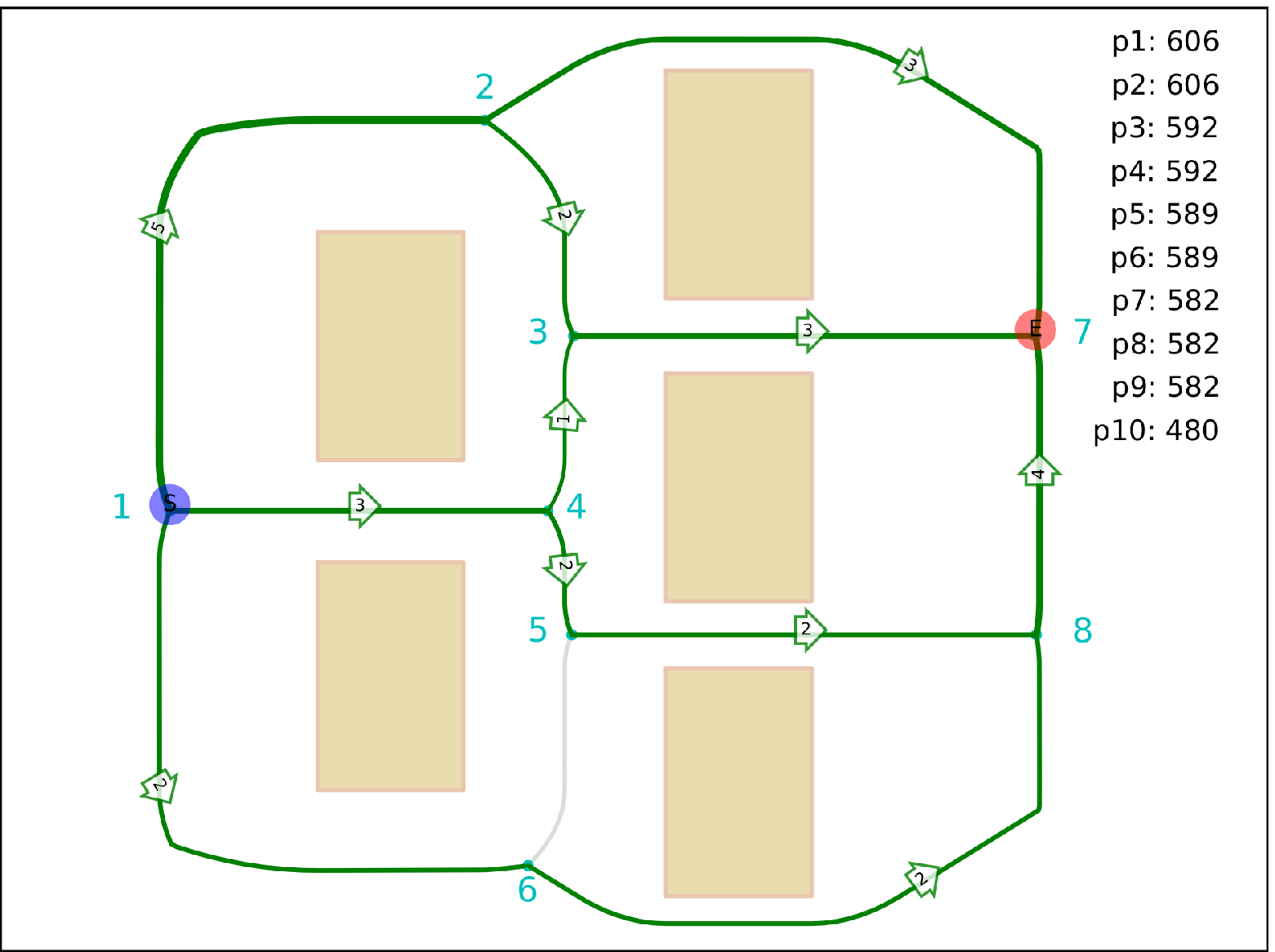}
  }
  \caption{The optimal solutions computed by the algorithm for a formation consisting of four robots in \protect\subref{subfig:simple_map_a}, and  ten robots formation in \protect\subref{subfig:simple_map_b}.}
  \label{fig:simple_map}
\end{figure}

\begin{figure}[htb]
  \centering
  \subfloat[]{ 
    \label{subfig:complexmaps_a}
    \includegraphics[width=0.48\textwidth]{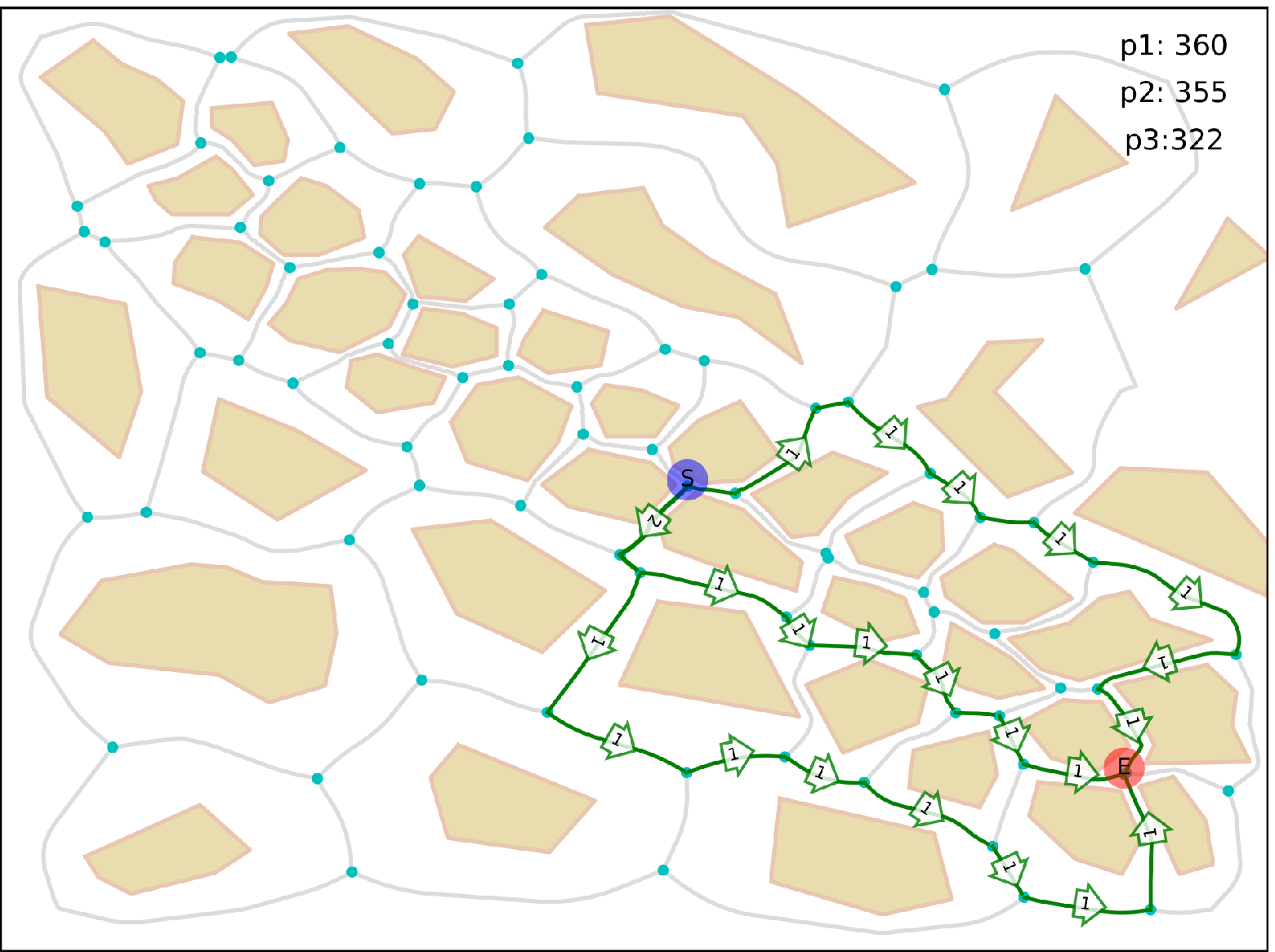}
  }\hfill
  \subfloat[]{ 
    \label{subfig:complexmaps_b}
    \includegraphics[width=0.48\textwidth]{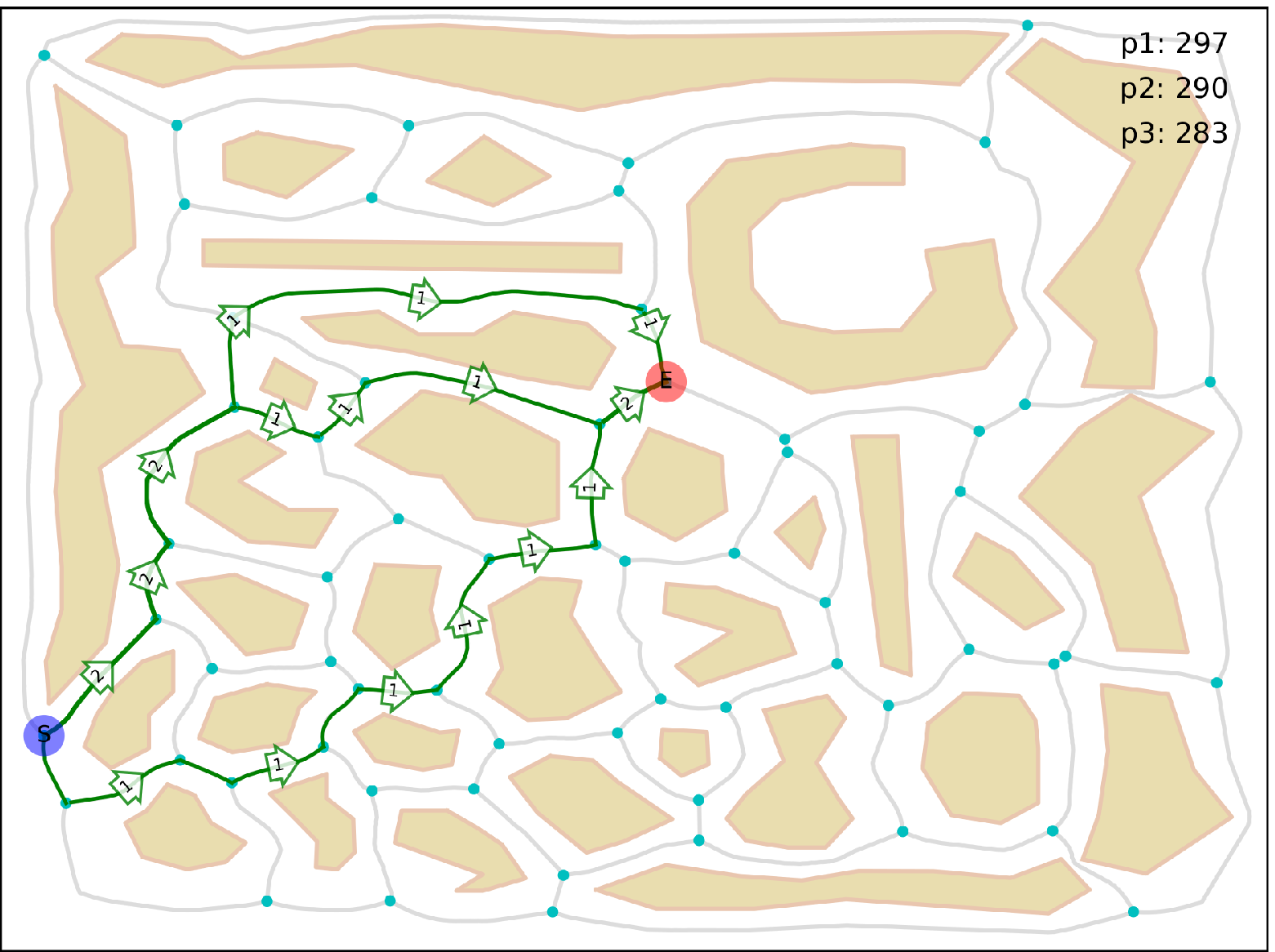}
  }\\
  \subfloat[]{ 
    \label{subfig:complexmaps_c}
    \includegraphics[width=0.48\textwidth]{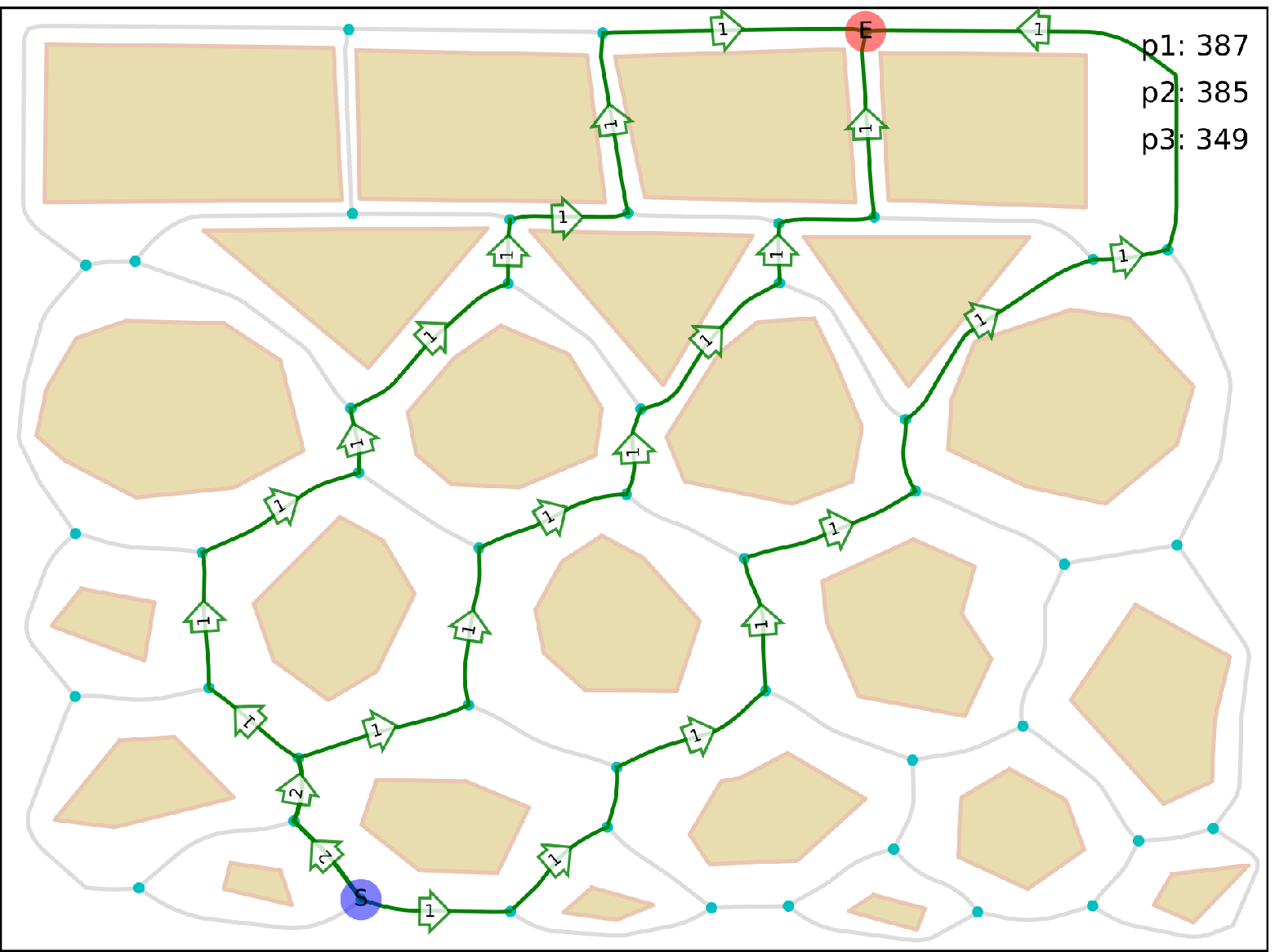}
  }\hfill
  \subfloat[]{ 
    \label{subfig:complexmaps_d}
    \includegraphics[width=0.48\textwidth]{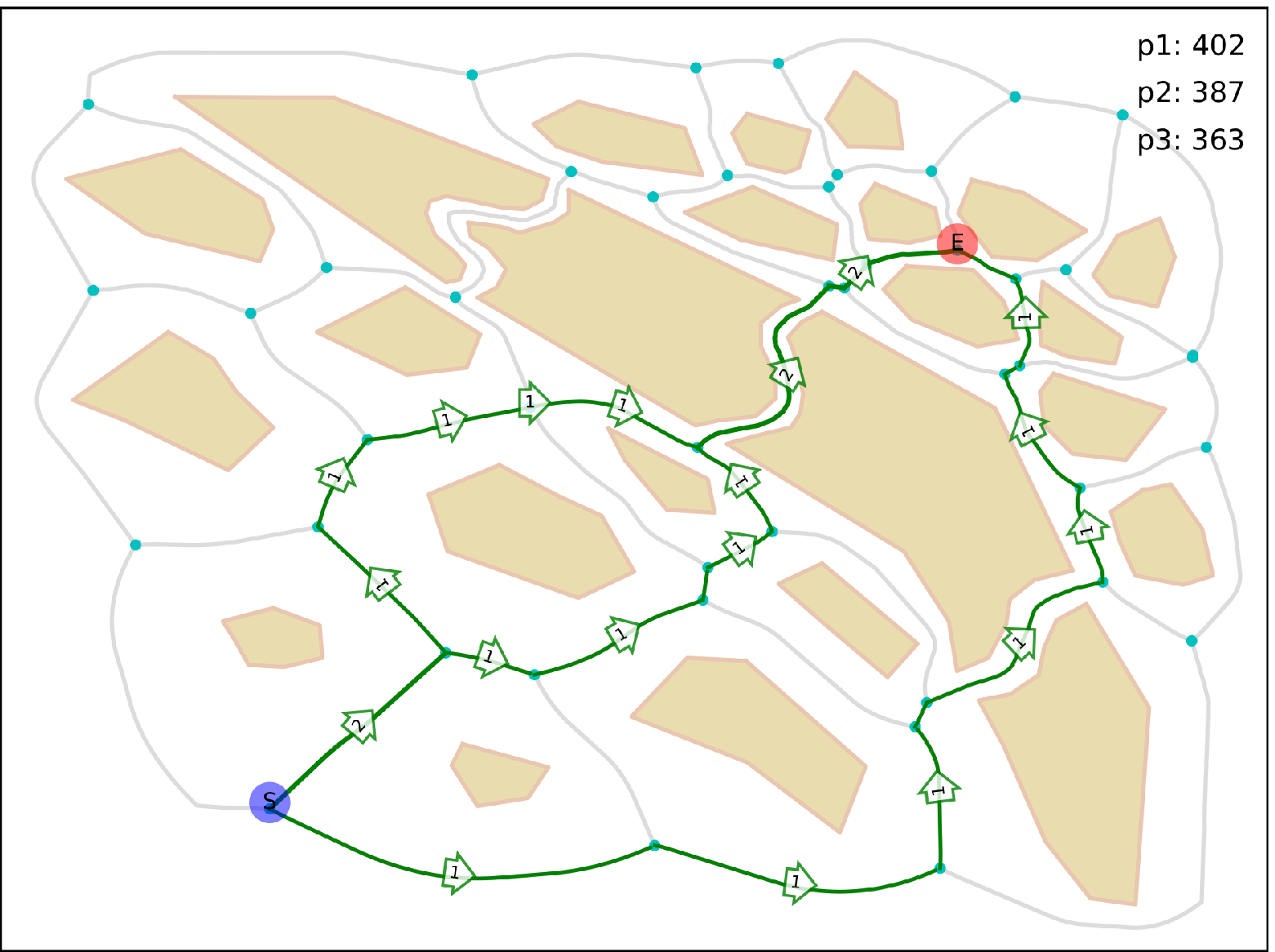}
  }
  \caption{The optimal solutions computed by the algorithm for a three robots formation in maps with various complexity.}
  \label{fig:complexmaps}
\end{figure}

In the Fig.~\ref{fig:simple_map} an example of a path planning in a simple map with $8$ vertices is shown.
For each edge of the graph a vector of costs is determined by the following equations,  
\begin{align*}
  e_{12} &= int(142.93 + r*19.99) &  e_{34} &= int(44.99 + r*31.24)\\
  e_{14} &= int(59.67 + r*38.43) &  e_{45} &= int(31.94 + r* 31.23)\\
  e_{16} &= int(142.88 + r*24.38) &  e_{58} &= int(73.37 + r*58.71)\\
  e_{27} &= int(153.37 + r*62.43) &  e_{56} &= int(59.92 + r*31.24)\\
  e_{23} &= int(58.41 + r*31.24) &  e_{68} &= int(141.78 + r*71.33)\\
  e_{37} &= int(72.96 + r*52.55) &  e_{78} &= int(76.048 + r*13.88)
\end{align*}
where $r\in {1, 2, \dots, R}$ is the number of robots travelling through the edge.
All the robots of the formation are initially at the blue start node, the formation splits and merges at different nodes, following the green paths.
The arrows in the graph represent the number of robots moving through each edge and the direction of the movement.
Fig.~\ref{subfig:simple_map_a} shows the solution for four robots. Note the number in square brackets drawn near each edge which express vectors of costs computed by the equations presented before for $r = 1$ to $r = 4$.
The optimal paths given by the algorithm are
\begin{align*}
  p_1 &= \{1, 2, 3, 7\}\quad &c_1 &= 449\\
  p_2 &= \{1, 4, 5, 8,7\}\quad &c_2 &= 420\\
  p_3 &= \{1, 2, 7\}\quad &c_3 &= 397\\
  p_4 &= \{1, 4, 3, 7\}\quad &c_4 &= 390,
\end{align*} 
and the formation cost is then given by the worst cost of the set, that is the cost of the path $p_1$, $c_1 = 449$.
In the Fig.~\ref{subfig:simple_map_b} the solution for ten robots formation is shown, in this case the solution found is
\begin{align*}
  p_1 &= \{1, 6, 8, 7\}    & c_1 &= 606  \quad & p_6    &= \{1, 4, 5, 8, 7\} & c_6    &= 589\\
  p_2 &= \{1, 6, 8, 7\}    & c_2 &= 606  \quad & p_7    &= \{1, 2, 7\}       & c_7    &= 582\\
  p_3 &= \{1, 2, 3, 7\}    & c_3 &= 592  \quad & p_8    &= \{1, 2, 7\}       & c_8    &= 582\\
  p_4 &= \{1, 2, 3, 7\}    & c_4 &= 592  \quad & p_9    &= \{1, 2, 7\}       & c_9    &= 582\\
  p_5 &= \{1, 4, 5, 8, 7\} & c_5 &= 589  \quad & p_{10} &= \{1, 4, 3, 7\}    & c_{10} &= 480
\end{align*}
Although the complexity of this map is relatively low, the problem becomes quickly very complex as the number of robots grows and therefore it is not easy to prove the optimality of the generated solutions.
In the same way, if the complexity of the map (i.e. the number of vertices) grows, the optimality of the solutions is very difficult to verify even with a low number of robots.
For example, Fig.~\ref{fig:complexmaps} shows a solution for a three robots formation in four complex maps, the map on \ref{subfig:complexmaps_a} has a total of $77$ nodes, the map on  \ref{subfig:complexmaps_b} has $62$ nodes and the maps on \ref{subfig:complexmaps_c} and \ref{subfig:complexmaps_d} have $45$ nodes.

\section{Conclusion}
\label{sec:conclusion}
A novel algorithm for multi-robot path planning was presented.
The algorithm finds a path for each robot of a formation, from a given node to all other nodes in the graph, considering the possibility of split and merge of the formation.
It is an extended version of the standard Dijkstra's algorithm, and it is therefore deterministic, complete and finds an optimal solution.
The algorithm is computationally heavy, and finding solutions for more than ten robots in maps with more than twenty nodes is very time consuming.

In the future work, improvements in realization of the constrains for formation merge will be considered, because this is the most time consuming part of the algorithm. Naturally, some aforementioned motion planning  approach for formations will be employed to form, together with the proposed solution, the integrated approach for formation planning. This will allow us to perform experiments in realistic setups in simulation and then with real robots.

\section*{ACKNOWLEDGMENT}
The work of M. Estefan\'ia Pereyra and R. Gast\'on Aragu\'as has been supported by the ``Multirrotores Aut\'onomos para Aplicaciones en Ambientes Exteriores'' project, U.T.N. PID UTI4534.
The work of Miroslav Kulich has been supported by European Community's HORIZON 2020 Programme under grant agreement No.~688117  ``SafeLog: Safe human-robot interaction in logistic applications for highly flexible warehouses''.%

\bibliographystyle{splncs03}
\bibliography{main}

\end{document}